\documentclass[times, review, 10pt]{elsarticle}

\usepackage{amssymb}
\usepackage{amsmath}
\usepackage{multirow}

\journal{Nuclear Physics B}

\begin{document}

\begin{frontmatter}



\title{GaitAdapt: Continual Learning for Evolving Gait Recognition} 


\author{Jingjie Wang}
\ead{23111492@bjtu.edu.cn}

\author{Shunli Zhang\corref{cor1}}
\ead{slzhang@bjtu.edu.cn}

\author{Xiang Wei}
\ead{xiangwei@bjtu.edu.cn}

\author{Senmao Tian}
\ead{23121588@bjtu.edu.cn}

\cortext[cor1]{Corresponding author}
\affiliation{organization={School of Software Engineering, Beijing Jiaotong University},
            addressline={}, 
            city={Beijing},
            postcode={100044}, 
            state={},
            country={China}}

\begin{abstract}
Current gait recognition methodologies generally necessitate retraining when encountering new datasets. Nevertheless,retrained models frequently encounter difficulties in preserving knowledge from previous datasets, leading to a significant decline in performance on earlier test sets. To tackle these challenges, we present a continual gait recognition task, termed GaitAdapt, which supports the progressive enhancement of gait recognition capabilities over time and is systematically categorized according to various evaluation scenarios. Additionally, we propose GaitAdapter, a non-replay continual learning approach for gait recognition. This approach integrates the GaitPartition Adaptive Knowledge (GPAK) module, employing graph neural networks to aggregate common gait patterns from current data into a repository constructed from graph vectors. Subsequently, this repository is used to improve the discriminability of gait features in new tasks, thereby enhancing the model’s ability to effectively recognize gait patterns. We also introduce a Euclidean Distance Stability Method (EDSN) based on negative pairs, which ensures that newly added gait samples from different classes maintain similar relative spatial distributions across both previous and current gait tasks, thereby alleviating the impact of task changes on the distinguishability of original domain features. Extensive evaluations demonstrate that GaitAdapter effectively retains gait knowledge acquired from diverse tasks, exhibiting markedly superior discriminative capability compared to alternative methods.
\end{abstract}




\end{frontmatter}


\begin{figure}[h!]
\centering
  \includegraphics[width=\linewidth]{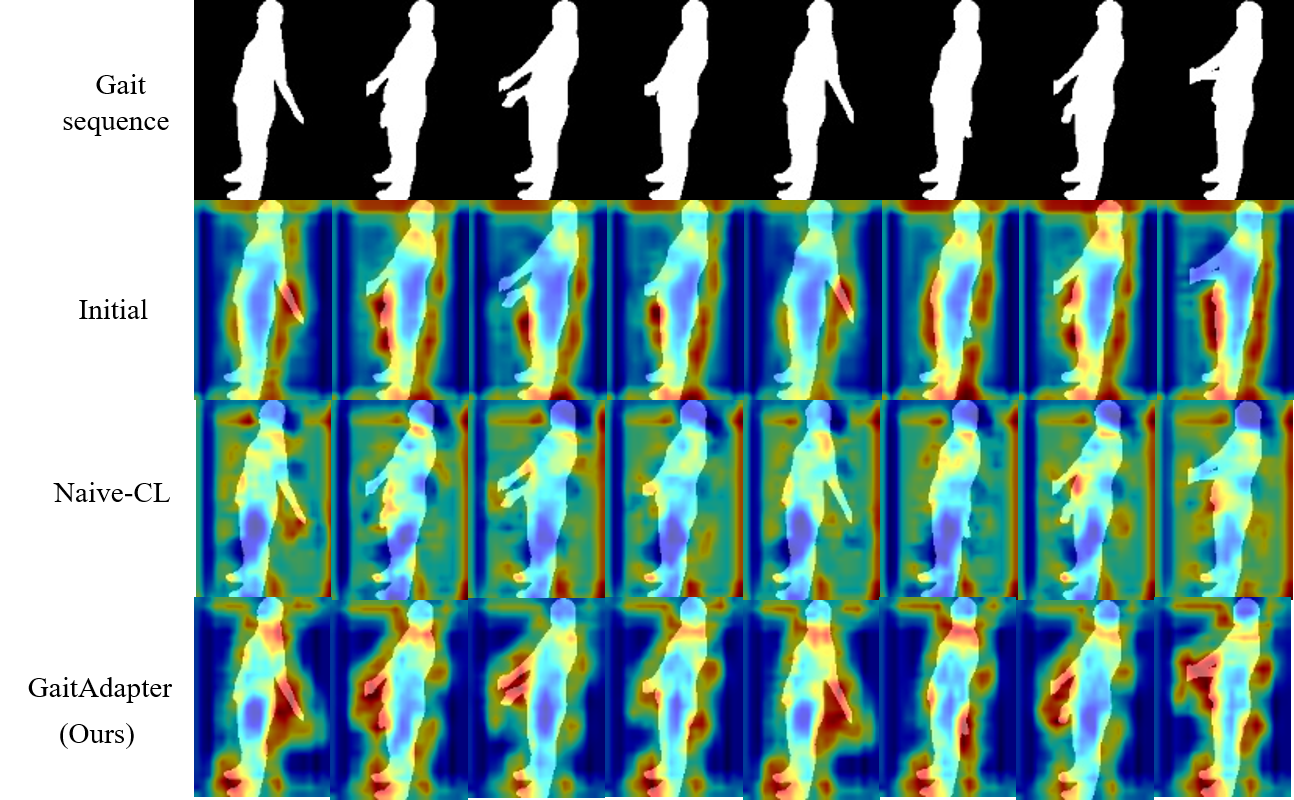}
  \caption{Feature Activation Heatmaps Across Learning Steps. The first row shows the original silhouette inputs. Subsequent rows visualize feature activations: (a) Without  continual learning Features, (b) Naive CL Features (traditional continual learning), and (c) GaitAdapter Features (our method). Heatmaps reveal how GaitAdapter preserves task-relevant activations while mitigating interference, compared to scattered patterns in naive CL and limited adaptability of without  continual learning features.}
  \label{f0}
\end{figure}
\section{Introduction}
Gait recognition, one of the most promising biometric recognition technologies, aims to identify and authenticate individual identity or behavior through non-intrusive methods. In recent years, gait recognition networks that integrate convolutional neural networks(CNN)\cite{hara2018can,he2016deep} or graph neural networks(GNN)\cite{zhou2020graph,wu2020comprehensive} have demonstrated robust gait representation capabilities\cite{fan2023opengait,chao2019gaitset,teepe2021gaitgraph,teepe2022towards}, though these are often confined to a predetermined sample domain\cite{yu2006framework}. In real-world recognition scenarios, gait recognition networks must possess the capacity for continual learning and adaptation to novel samples. For instance, urban surveillance cameras routinely capture pedestrian gait images that include diverse identities, viewpoints, attires, and motion patterns. A key challenge in gait recognition is the dynamic expansion of the discernible domain.
\begin{figure}[h]
  \centering
   \includegraphics[width=0.8\textwidth]{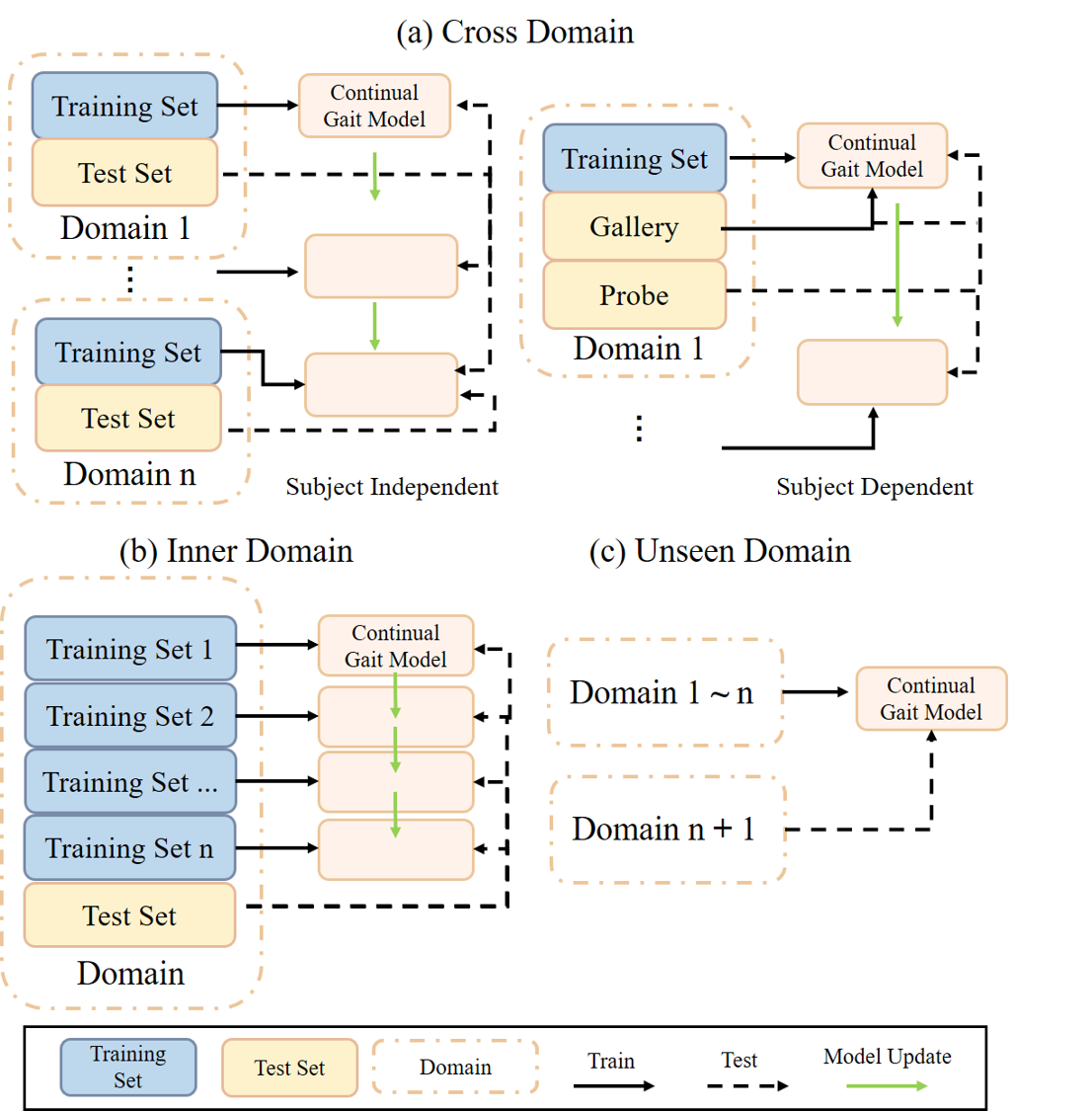}
\caption{Three scenarios in GaitAdapt. (a)The model continually trains on varied sections of a single domain and evaluates using the test set within the domain. (b) Cross domain. The model trains on diverse domain data and evaluates the test set from the learned domain. (c)Unseen Domain. The evaluation uses the test set from the unlearned domain.
}
    \label{f1}
\end{figure}

To address the aforementioned challenges, we propose a novel gait recognition task tailored for practical applications, named GaitAdapt. This task requires the model to learn and recognize human gait across a variety of continual scenarios, Based on the different gait recognition scenarios, we categorize GaitAdapt into three distinct cases, as illustrated in Figure \ref{f1}. GaitAdapt differs significantly from conventional gait recognition tasks. Firstly, most prevailing gait recognition methodologies rely on silhouette image sequences and 2D/3D skeletons\cite{mao2020gait,stenum2021two}, where the information is primarily focused on contour edges and joint points, resulting in low information entropy within the samples. Consequently, GaitAdapt is a fine-grained continual recognition task, in contrast to other continual biometric recognition techniques that rely on RGB images\cite{wang2021deep,jiang2023cross}. Secondly, existing gait recognition methods typically construct feature representation spaces using metric learning. In fine-grained continual recognition tasks, the original feature representation space tends to be more unstable, making it challenging for purely regularization-based continual learning methods to be effective. Finally, GaitAdapt is a temporal-related task compared to traditional continual learning tasks.

We present the GaitAdapter network to tackle the GaitAdapt task, focusing on continual scenarios without data replay. The core of GaitAdapter lies in the GaitPartition Adaptive Knowledge (GPAK) module, functioning as a repository that incrementally accumulates and refines fine-grained gait knowledge through graph convolutional networks. Prior studies\cite{fan2020gaitpart} have confirmed the effectiveness of part-based methods, which are capable of preserving more local information and fostering stronger associations among local details than global features. Within GaitAdapter, the bipartite-graph convolution method further facilitates the interaction between the current part-level knowledge and the accumulated knowledge stored in the GPAK module, enabling the network to integrate general gait knowledge obtained from previous training stages. Furthermore, GaitAdapter employs a Euclidean distance stability method based on negative sample pairs, which preserves the consistency of spatial relationships within gait feature sets across various training stages, thereby mitigating the disruptive effects of newly learned features on the original spatial distribution.

As shown in Figure \ref{f0}, after several steps of training in the gait model, we backtesting the earliest learned sample. Without using continuous learning methods (W/O CL), the model's region of interest showed a significant shift and focused more on the irrelevant area at the top of the image, indicating a serious forgetting phenomenon in the model. After using Native continuous learning methods (Native CL), this phenomenon was not alleviated. We believe that the possible reason is that, unlike traditional image classification or object detection, gait recognition relies more on fine-grained features (such as angles and local motion patterns), and traditional CL methods do not consider the importance changes of local features in gait data (such as the criticality of leg movements), resulting in the model being dominated by irrelevant global features (such as background) in the incremental learning process. Our method introduces the GPAK module and EDSN method, allowing the model to focus on the effective area even after several training steps.

The contributions of this paper are summarized as follows:
\begin{itemize}
\item{} Proposal of a continual gait recognition task, GaitAdapt, is tailored for real-world applications. GaitAdapt explores the ability of gait models to learn and adapt to different identities and scenarios in continual learning.
\item{} Introduction of GaitAdapter, which enhances gait recognition without relying on data replay, by incorporating mechanisms for continual knowledge accumulation and stability. 
\item{} Integration of various continual learning scenarios and the proposal of a comprehensive evaluation protocol for GaitAdapt. Our approach reveals considerable potential for advancing the field of continual gait recognition.
\end{itemize}

\section{Related Work}
\subsection{Gait Recognition}
Gait refers to the posture of walking, a biological behavioral characteristic perceivable at a distance. Gait recognition technology identifies individuals by distinguishing their walking patterns. Compared to other biometric recognition techniques, such as facial recognition, gait recognition offers non-contact, non-invasive, and difficult-to-disguise features. Utilizing deep learning techniques to construct gait recognition networks is the current mainstream approach in gait recognition. Classical gait networks such as GaitSet\cite{chao2019gaitset}, GaitPart\cite{fan2020gaitpart}, GaitBase\cite{fan2023opengait}, GaitGL \cite{lin2022gaitgl}, and DyGait\cite{wang2023dygait} all employ 2D/3D convolutional neural networks for global or local feature extraction. Additionally, successes have been achieved with GaitGraph \cite{teepe2021gaitgraph} and GaitGraph2\cite{teepe2022towards} based on graph neural network technology, as well as SwinGait\cite{fan2023exploring} based on visual Transformer.

The collection of gait samples is another focus in the field of gait recognition. In the early stages, gait samples compressed gait information into single images, such as gait energy images(GEI)\cite{shiraga2016geinet}. However, compressing images often leads to loss of spatio-temporal information, making it impossible to model fine-grained spatio-temporal information. Therefore, researchers began using gait silhouette sequences\cite{chao2019gaitset} as model inputs, becoming the primary choice for current gait sample forms. Furthermore, efforts have been made to construct datasets containing richer representations of gait information to overcome the instability of single features and enable gait models to operate in non-laboratory environments. Examples include the Gait3D\cite{zheng2022gait} dataset utilizing SMPL\cite{loper2023smpl}, the Gait3D-Parsing\cite{zheng2023parsing} dataset employing human-parsing techniques, and the CCPG\cite{li2023depth} dataset for clothing change scenarios.

Although progress has been made in gait recognition, gait models trained on single and straightforward scenes cannot guarantee generalization and scalability in dynamically changing and complex scenarios. We are committed to enhancing the representation capacity of gait models in continual scenes.
\subsection{Continual learning}
Continual learning means that the model needs to continually receive new data and adapt to it while maintaining the effectiveness and utilization of the knowledge learned in the past. Continual learning typically involves some challenges, such as Catastrophic Forgetting and Sample Efficiency. Common continual learning methods can be divided into three categories: parameter isolation, regularization-based, and feature replay-based\cite{de2021continual}.

Parameter isolation methods aim to minimize parameter overlap and interference between tasks by learning separate network parameters for each task, avoiding deviations from previous optimal solutions. However, while dynamically increasing network parameters is common, it is impractical for long incremental tasks or when storage space is limited. Replay-based continual learning overcomes forgetting by saving a set of examples from each task\cite{hou2019learning,hayes2020remind}, but in biometric recognition, the practice of saving examples may bring privacy risks. In regularization-based continual learning, knowledge distillation is a widely used regularization method that reduces forgetting by aligning features\cite{liu2023lifelong,chenshen2018memory} or aligning prediction probabilities\cite{li2017learning}. Currently, continual learning methods based on knowledge distillation and other auxiliary techniques have achieved good results in some identity feature recognition fields, such as person re-identification\cite{pu2021lifelong}.

However, more than continual methods based on pure knowledge distillation are needed for GaitAdapt. 1) In GaitAdapt, the number of recognizable individuals is much larger than in traditional image classification tasks; for example, their popular benchmarks include MNIST\cite{lecun1998gradient}, CIFAR-10/100\cite{krizhevsky2009learning}, and ImageNet\cite{russakovsky2015imagenet}. In contrast, the number of individuals to be handled by gait recognition models in real scenarios is unknown. 2) GaitAdapt is similar to fine-grained retrieval tasks\cite{chen2020exploration}. The interclass sample appearance variation in GaitAdapt is much more subtle than in typical classification tasks and biometric recognition tasks, making GaitAdapt more challenging. 3) Existing gait recognition methods typically construct feature representation spaces through metric learning. In fine-grained continual recognition tasks, the original feature representation space is even more unstable, making traditional continual learning methods less effective.
\begin{figure}[h!]
  \centering
   \includegraphics[width=\textwidth]{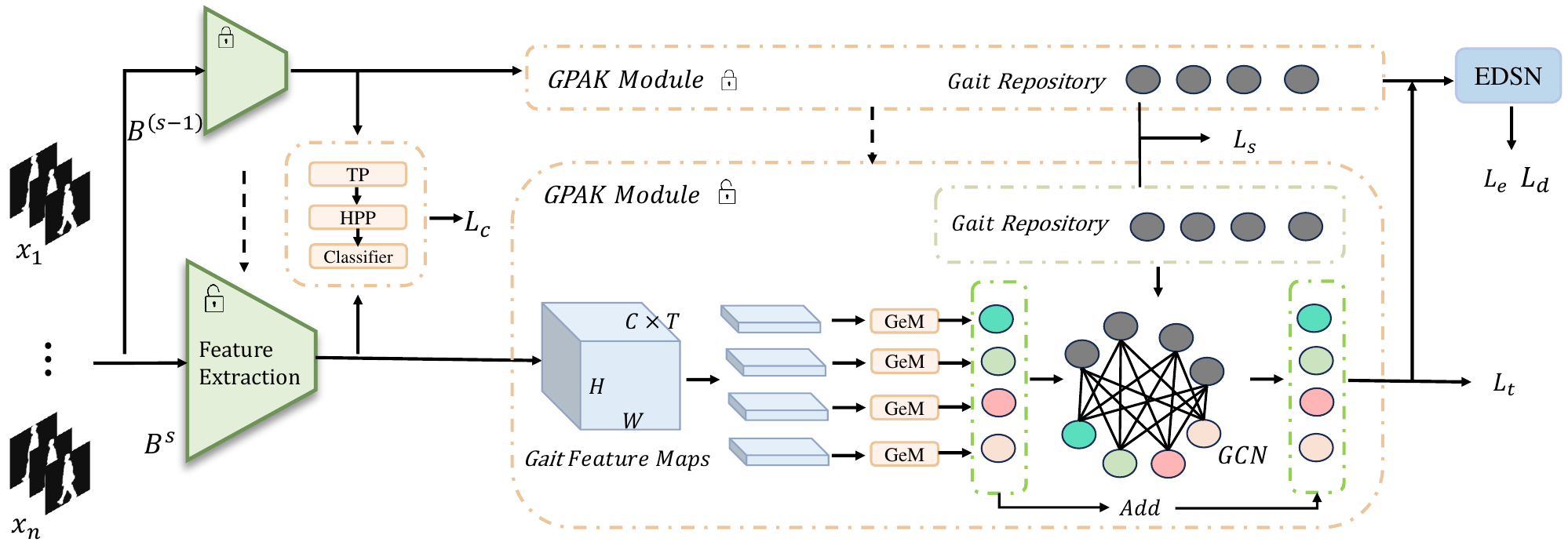}
  \caption{Overview of the proposed GaitAdapter. 
Solid-line arrows indicate the flow direction of model features, while dashed-line arrows represent the replication of model parameters.
}
\label{f2}
\end{figure}

\section{GaitAdapter Without Data Replay}
In this section, we present the specific details of the proposed GaitAdapter. GaitAdapter implementation is mainly accomplished through two parts: the GaitPartition Adaptive Knowledge module, and the Euclidean Distance Stability Method. The overall model structure is shown in Figure \ref{f2}. We aim to iteratively improve the performance of the model through continuous training steps, allowing it to handle complex tasks or environmental changes without relying on data replay. In the following subsections, each subprocess is elaborated upon.
\subsection{Problem Deﬁnition and Formulation}
As for GaitAdapt, the training and testing process needs to be divided into $S$ steps, learning gait knowledge in $T$ domains, $T<=S$. Suppose we have a data stream $\mathcal{D} =\left \{D^{(s)}  \right \}^{S}_{s=1} $,the dataset for the $s$-th step $D^{(s)}=\{D^{(s)}_{tr}, D^{(s)}_{te}\}$ can originate from any domain, where $D^{(s)}_{tr}$ represents the training set and $D^{(s)}_{te}$ represents the test set, $D^{(s)}_{tr}\bigcap D^{(s)}_{te} = \emptyset$. In addition, $D^{(s)}_{te} = \{(g_i, p_i)\}_{i=1}^{\left |D^{(s)}_{te}  \right | }$, where $g$ represents the gallery set, and $p$ represents the probe set, and $p\cap D_{tr} = \emptyset$. GaitAdapt strictly adheres to the continual process and does not utilize data from preceding steps during the training phase. We employ various strategies to evaluate all test sets $D_{te}$, as shown in the next section.

\subsection{Feature extractor with extensible classification header}
We introduced a multilayer convolutional neural network\cite{fan2023opengait}, as the feature extraction network for the GaitAdapt task due to its favorable transfer ability. The network is denoted as $B(\cdot, \theta)$, where $\theta$ represents the parameters of the extraction network. In order to accommodate the continual expansion of the number of individual identities, we appended a sustainable extensible classification header $c(\cdot, \phi)$ to the end of the $B$ network, where $\phi$ denotes the classification head parameters. Thus, the network output for a gait sample $x$ can be expressed as $c(B(x, \theta), \phi)$. In traditional gait recognition tasks, cross-entropy is employed to compute the loss between the network output and the actual person label $y$, as follows
\begin{equation}
L^s_{c} = -\frac{1}{N^s} \sum_{i}^{} \sum_{j=1}^{C^s} y_{ij}log(c_j(B(x_i, \theta), \phi))
\end{equation}
where $L^s_c$ denotes the supervised loss at step $s$, $N^s$ represents the number of training samples at step $s$, and $C^s$ indicates the number of individual identities at step $s$, $i$ represents the $i$-th sample and $j$ represents the $j$-th personnel identification.

\subsection{GaitPartition Adaptive Knowledge}
The GaitPartition Adaptive Knowledge (GPAK) module focuses on capturing meaningful and discriminative patterns from current gait features, storing them in specialized repositories for subsequent training steps. Drawing inspiration from the parallels between deep neural networks and human memory processes, graph neural networks are utilized as the medium for accumulating cross-step gait knowledge. Specifically, the GPAK module operates through four key processes: establishing the gait repository, constructing a part-based gait knowledge graph, transferring learned knowledge, and injecting and maintaining accumulated knowledge.

\textbf{GaitPartition Knowledge Graph.} Previous studies have shown that part-based approaches can enable models to learn richer gait patterns. Therefore, we horizontally split the global gait feature $\mathcal{F} = B(x, \theta)$ extracted by the feature network $ B(., \theta)$ to obtain $ m $ local-level features $F' \in \mathbb{R} ^{m\times S \times C\times T \times \frac{H}{m}  \times W}$, where $m$ represents the number of parts, $S$ represents batch size, $C$ represents the feature dimension, $T$ represents the temporal length and $H, W$ represent the size of the global feature map. Then, to enable our method to adaptively integrate spatiotemporal information, we employ Generalized-Mean pooling \cite{wang2018selective} on $ F'$ to obtain $f \in \mathbb{R} ^{mS \times C}$, shown as follows
\begin{equation}
f =(GP_{avg}(F')^{\alpha })^{\frac{1}{\alpha } }
\end{equation}
where $GP_{avg}$ represents average pooling, where $\alpha$ is a learnable parameter. When  $a \to \infty$ , average pooling transitions to max pooling.
Then, we treat $f$ as graph vectors and use a edgeless graph $G^f(A^f, f)$  to establish knowledge associations among all part-level features, where $A^f = 0$ is the adjacency matrix of graph $G^f$.By maintaining $A^f = 0$ throughout, we enforce complete feature isolation, preventing any interference between part-level representations while preserving their intrinsic discriminative properties. This absolute decoupling is particularly meaningful when the compositional semantics naturally reside in individual features rather than their interactions.

Moreover, like $G_f$, we also employ a edgeless graph $G_r(A^r,K)$ to serve as the repository, where $K$ represents the vertex set, a parameter set obtained through random initialization for learning, and $A^r = 0$ denotes the adjacency matrix of the vertex set. The crucial point to note is that, although the construction methods of $G^f$ and $G^r$ are similar, they fundamentally differ in their internal knowledge construction and roles. $G^f$ is responsible for constructing part-based gait knowledge during the current training step, while $G^r$ is responsible for maintaining past general gait knowledge and memorizing gait knowledge from the current training step.

\textbf{Gait knowledge transfer.} 
Gait knowledge transfer leverages bipartite-graph convolution to connect and transfer knowledge between the $G_f$ and $G^r$, we construct a fully connected graph containing $G_r$ and $G_f$, referred to as the transfer graph  $G_t$. Since $G_t$ contains all vertex from both graphs, we represent the vertex set $V^t$ in $G_t$ as
\begin{equation}
V^t = \left \{v_i | v_i \in f \text{ or } v_i \in K \right \}
\end{equation}

Furthermore, given $f_i$ as the $i$-th  vertex in $G_f$ and $K_j$ as the $j$-th vertex in $G_r$, the edges connected by vertex are defined as
\begin{equation}
A^c_{ij} = \frac{exp(-\frac{1}{2} \left \| f_i-K_j \right \|^2_2 )}{ {\textstyle \sum_{k=1}^{N^r}}  exp(-\frac{1}{2} \left \| f_i-K_k \right \|^2_2 )}.
\end{equation}
It should be noted that unlike $G_r$ and $G_f$, the weights between vertex across graphs possess symmetry, i.e.,  $A^c_{ij} = A^c_{ji}$. Based on this, we can construct the adjacency matrix $A^t$ of $G_t$ as follows
\begin{equation}
A^t =\begin{bmatrix}
 0 & A^c\\
 (A^c)^T & 0
\end{bmatrix}.
\end{equation}

Finally, we accomplish the transfer of disparate structural gait knowledge through graph convolution operations\cite{kipf2016semi}, which can be formulated as
\begin{equation}
V^F = \varrho A^t(V^tW^t) 
\end{equation}
where $V^F \in \mathbb{R}^{(mS + N^r) \times C}$, $mS$ denotes the number of vertex in $G_f$, $N^r$ denotes the number of vertex in $G_r$. $\varrho$ represents the activation function, and $W^t$ represents the learnable weights of the graph convolution. At this point, the part-based features have undergone knowledge transfer from the knowledge repository, resulting in vertex features $V^f =\left\{ V^F_i | i \in [1, mS] \right\} $.

\textbf{Gait Knowledge Injection \& Maintenance.} To enhance the training efficiency and performance of the gait model, we integrate the representations extracted by the feature network with the representations injected by the GPAK module, where $F$ serves as the representation set for discriminating gait identities,
\begin{equation}
F = V^f + f 
\end{equation}

Additionally, we utilize a stability loss to constrain the excessive updates to the repository. The stability loss $L_s$ is defined as
\begin{equation}
L_s = \frac{1}{N^r} \sum_{i=1}^{N^r} ln(1+exp(\triangle(K_i,\widetilde{K} _i) ))
\end{equation}
where $\widetilde{K}$ represents the repository vertex set of the last training step, and $N^r$ indicates the number of vertex in the repository,$\triangle$ denotes the measure of similarity between vectors. We achieve the maintenance of the repository by imposing consistency regularization on each vertex between new and old tasks. 
\begin{figure}[h]
  \centering
   \includegraphics[width=\linewidth]{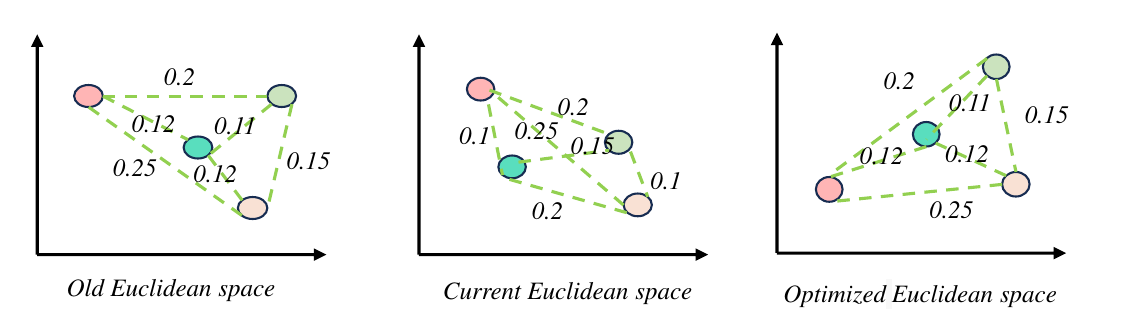}
  \caption{Illustration of the Euclidean Distance Stability Method, where different colors represent gait features of different identities, and numbers denote relative distances. The optimized spatial distribution can enhance the stability of training across different steps.
}
\label{f3}
\vspace*{-1em}
\end{figure}
\subsection{Euclidean Distance Stability}
Gait recognition typically discriminates between positive and negative sample pairs by learning similarity measures of gait features. Due to the subtle differences between gait samples (compared to image features or other biological characteristics), traditional continual learning methods applied to conventional classification tasks are challenging to apply to GaitAdapt directly. Rather than focusing on person-category mappings, GaitAdapt should pay more attention to the Euclidean spatial distribution of the extracted features, which directly determines gait recognition performance. Therefore, we employ the Euclidean distance stability method based on negative pairs(EDSN) to constrain the differences in feature space distribution across tasks. The method is illustrated in Figure \ref{f3}.

Given $F(s)$, which is the representation extracted by GPAK from the gait sample $x$ in the $s$-th step training of GaitAdapter, and $F(s-1)$ denotes the representation obtained when the input current sample into the gait recognition model at the $(s-1)$-th step. As the gait feature extraction model and GPAK module continue to update,  $x$ may not have the same spatial distribution across different model versions, which can be prone to catastrophic forgetting in the continual gait recognition model. We calculate the relative positions between sample pairs in $F(s)$  and  $F(s-1)$, denoted as $d(s)$ and $d(s-1)$, respectively. The calculation of $d$ is as follows
\begin{equation}
d_{ij} = \frac{exp(-\frac{1}{2} \left \| F_i-F_j \right \|^2_2 )}{ {\textstyle \sum_{k=1}^{N_i^n}}  exp(-\frac{1}{2} \left \| F_i-F_k \right \|^2_2 )} .
\end{equation}

It should be noted that to assist GaitAdapter in capturing boundary information between positive and negative sample pairs more effectively, we exclude all relative positions among positive sample pairs, specifically when the identities are different $C_i \ne C_j$, where $N_i^n$ denotes the number of negative samples for the $i$-th sample. Subsequently, we utilize a stability loss to facilitate the model in learning the relative spatial representation of the $s$-th step samples on the $(s-1)$-th step model, thereby preventing interference from model updates on the existing representation space. The stability loss $L_e$ can be expressed as
\begin{equation}
L_e = \sum_{i,j}^{} d(s-1)_{ij}  log\frac{d(s-1)_{ij}}{d(s)_{ij}} .
\end{equation}

While $L_e$ effectively mitigates feature space drift, the model's decision layer output (i.e., Logit distribution) can still exhibit considerable variations, potentially impairing the model's capacity to accurately classify previously learned task categories.Incorporating Logit distillation loss introduces supplementary constraints at the output layer, ensuring the model preserves its discriminative capacity for previously learned categories while adapting to new tasks. Furthermore, we adopted traditional continual learning approaches\cite{li2017learning} to regularize the classification results from different training steps, aiming to optimize the class space distribution across various steps. This regularization loss is defined as
\begin{equation}
L^s_{d} = -\frac{1}{N^s} \sum_{i}^{} \sum_{j=1}^{C^{\hat{s}}}c_j(B(x_i, \hat{ \theta}), \hat{\phi})log(c_j(B(x_i, \theta), \phi))
\end{equation}
where $C^{\hat{s}}$ indicates the number of individual identities before current step, and $\hat{ \theta}$, $\hat{\phi}$ represent the network parameters from the last training step, $i$ represents the $i$-th sample and $j$ represents the $j$-th personnel identification.

\subsection{Optimization}
GaitAdapter follows the prevalent approach in existing gait recognition models by employing a hybrid training strategy, as commonly observed in BoT (Bag of Tricks)\cite{luo2019bag}, combining the usage of ID Loss and Triplet Loss. The ID Loss typically involves cross-entropy computation in Section 3.2. Furthermore, Triplet Loss is utilized to learn the similarity metric among gait samples. Its main idea is to train the model by constructing triplet samples comprising an anchor sample ($F_a$), a positive sample ($F_p$), and a negative sample ($F_n$). The objective of the loss function is to minimize the distance between the anchor sample and the positive sample while maximizing the distance between the anchor sample and the negative sample. Triplet loss\cite{hermans2017defense} is defined as follows
\begin{equation}
L_t = \frac{1}{N_T} \sum_{a,p,n}^{} ln(1+exp(\triangle (F_a, F_p) - \triangle (F_a, F_n))
\end{equation}
where $N_T$ represents batch size of triplet samples.

In the aforementioned, during the $s$-th training step, we utilize the training dataset $D^{(s)}_{tr}$ to train the entire model and optimize all trainable parameters using the comprehensive loss function $L^s$, which can be denoted as
\begin{equation}
L^s= L^s_c + L_t + L^s_d + L_s + L_e
\end{equation}

\begin{table*}[htbp]
 \caption{The dataset statistics used in GaitAdapt. ‘*/10’ denotes that the identity set is divided into 10 parts, and “(+n)” represents merging n identities from the gallery into the training set.‘-’ indicates that the dataset was not used.The "in-dpt" represents a subject-independent evaluation protocol, while "dpt" indicates a subject-dependent evaluation protocol.
}
\resizebox{\textwidth}{35mm}{
\begin{tabular}{c|c|c|c|ccc|ccc}
\hline
\multirow{2}{*}{\textbf{Datasets Name}} & \multirow{2}{*}{\textbf{Scale}} & \multirow{2}{*}{\textbf{Type}} & \multirow{2}{*}{\textbf{Evaluation Protocol}} & \multicolumn{3}{c|}{\textbf{Original Identities}}                     & \multicolumn{3}{c}{\textbf{Selected Identities}}   \\ \cline{5-10} 
                                        &                                 &                                &                                             & \textbf{Train}        & \textbf{Query}        & \textbf{Gallery}      & \textbf{Train} & \textbf{Query} & \textbf{Gallery} \\ \hline
\multirow{3}{*}{CASIA-B}          & \multirow{3}{*}{small}          & \multirow{3}{*}{Constrained}   & Inner Domain                                & \multirow{3}{*}{74}   & \multirow{3}{*}{50}   & \multirow{3}{*}{50}   & -              & -              & -                \\
                                        &                                 &                                & Cross Domain(In-dpt)                                &                       &                       &                       & 74             & 50             & 50               \\
                                        &                                 &                                & Cross Domain(dpt)                           &                       &                       &                       & 74 (+50)       & 50             & 50               \\
\multirow{3}{*}{OUMVLP}                 & \multirow{3}{*}{large}          & \multirow{3}{*}{Constrained}   & Inner Domain                                & \multirow{3}{*}{5153} & \multirow{3}{*}{5154} & \multirow{3}{*}{5154} & 5153/10        & 5154           & 5154             \\
                                        &                                 &                                & Cross Domain(In-dpt)                   &                       &                       &                       & 1000           & 5154           & 5154             \\
                                        &                                 &                                & Cross Domain(dpt)                           &                       &                       &                       & 5153 (+5154)   & 5154           & 5154             \\
\multirow{3}{*}{CASIA-E}                & \multirow{3}{*}{large}          & \multirow{3}{*}{Constrained}   & Inner Domain                                & \multirow{3}{*}{507}  & \multirow{3}{*}{507}  & \multirow{3}{*}{507}  & 200/10         & 300            & 300              \\
                                        &                                 &                                & Cross Domain(In-dpt)                     &                       &                       &                       & 200            & 300            & 300              \\
                                        &                                 &                                & Cross Domain(dpt)                 &                       &                       &                       & 200 (+300)     & 300            & 300              \\
\multirow{3}{*}{Gait3D}                 & \multirow{3}{*}{mid}            & \multirow{3}{*}{Unconstrained} & Inner Domain                                & \multirow{3}{*}{3000} & \multirow{3}{*}{1000} & \multirow{3}{*}{1000} & 3000/10        & 1000           & 1000             \\
                                        &                                 &                                & Cross Domain(In-dpt)                     &                       &                       &                       & 1000           & 1000           & 1000             \\
                                        &                                 &                                & Cross Domain(dpt)                            &                       &                       &                       & 1000 (+1000)   & 1000           & 1000             \\ \hline
GREW                                    & mid                             & Unconstrained                  & \multirow{3}{*}{Unseen Domain}              & 20000                 & 6000                  & 6000                  & -              & 6000           & 6000             \\
SUStech-1K                              & mid                             & Unconstrained                  &                                             & 500                   & 500                   & 500                   & -              & 500            & 500              \\
CCPG                                    & large                           & Unconstrained                  &                                             & 100                   & 100                   & 100                   & -              & 100            & 100              \\ \hline
\end{tabular}}
\label{t1}
\end{table*}

\section{Experiments}
\subsection{Continual Evaluation Protocol}
We introduce the GaitAdapt evaluation protocol to comprehensively assess the performance of gait recognition models in continual scenarios and provide corresponding evaluation methods for future researchers. Depending on the different continual learning scenarios (see Figure \ref{f1}), the GaitAdapt evaluation protocol can be divided into the following three types:
\begin{itemize}
\item{}\textbf{Cross-domain Protocol}: This protocol evaluates continual learning under domain shifts, assessing both knowledge retention and acquisition. It employs subject-dependent and subject-independent configurations\cite{sepas2022deep}, differentiated by gallery set inclusion during training. Crucially, the probe set never overlaps training data to avoid bias. Real-world deployments inherently involve dynamic domains, where galleries become integral to continual model optimization, justifying their inclusion in the learning process.
\item{}\textbf{Inner-domain Protocol}: Evaluating the performance of gait models in inner-domain scenarios. Inner-domain means that samples are collected under identical environments and conditions. Inner domain evaluation can demonstrate the adaptability of continual learning methods to specific scenarios.

\item{}\textbf{Unseen-domain Protocol}: This protocol allows models trained in visible domains to be migrated for evaluation in unknown scenarios. Continual gait learning models should possess generalization capabilities to cope with sudden changes in data distribution due to scene transitions.

\end{itemize}
According to the protocol, we reorganized the gait dataset for experimentation. As shown in Table \ref{t1}.

\subsection{Datasets Details}
 According to the protocol, we collected typical gait datasets and categorized them into two classes, one class is used for training and testing in the visible domain (including inner-domain, cross-domain, and subject-dependent), and the other is used for model generalization testing in the unseen domain. CASIA-B\cite{yu2006framework}, CASIA-E\cite{song2022casia}, OUMVLP\cite{takemura2018multi}, and Gait3D\cite{loper2023smpl} are used in the visible domain. These four datasets have entirely different scales, and there are significant differences in gait patterns, especially Gait3D, which originates from unconstrained environments, adding to the difficulty of GaitAdapt cross-domain learning. Secondly, to balance the differences in the scale of different datasets, we randomly selected a subset of identities for training for some larger datasets. Furthermore, we divided the randomly selected identities into ten sub-datasets in the inner domain to meet the training requirements, Notably, due to the limited sample size of the CASIA-B dataset, it was excluded from our inner-domain experiments. Finally, in the Subject-dependent task, we treated the gait gallery as part of the training set for training and evaluated the model.
 
Compared to the visible domain, the Unseen datasets contains more complex and realistic gait samples, such as GREW\cite{zhu2021gait} and SUStech-1K\cite{shen2023lidargait}, which performed outstandingly in gait recognition competitions, while the CCPG\cite{li2023depth} dataset contains rich disguise information. The complex and variable unseen gait patterns will further validate the representation ability of the continual gait model.

\subsection{Baseline Details}
We investigated methods in traditional continual learning and other biometric recognition domains and replicated these methods in the gait recognition task. The methods used for comparison include sequential fine-tuning (SFT), Learning without Forgetting (LwF)\cite{li2017learning}. Similarity-Preserving Distillation (SPD)\cite{tung2019similarity}. Continual Representation Learning (CRL)\cite{zhao2021continual}. Adaptive Knowledge Accumulation (AKA)\cite{pu2021lifelong}. The methods outlined below are all non-replay-based continual learning approaches, and replay-based continual learning methods are not included in the scope of comparison.
\begin{itemize}
    \item \textbf{Sequential Fine-Tuning (SFT)}: is a method where a model is incrementally trained on multiple tasks in a sequence. In this approach, the model is first trained on the data from the initial task. After this initial training, the model's parameters are fine-tuned on the next task in the sequence, using the weights from the previous task as initialization. This process continues for all subsequent tasks. 
    \item \textbf{Learning without Forgetting (LwF)}: is a technique for implementing continual learning in neural networks, aimed at training new tasks while preserving the performance on previously learned tasks. The method initializes the network with shared parameters $\theta_s$ and task-specific parameters $\theta_o$ for each old task, while using the new task data $X_n$ to generate the old task output $Y_o$ through a pre-trained network. New task parameters $\theta_n$ are then randomly initialized. During training, the method defines the old task predictions $\hat{Y_o}$ and new task predictions $\hat{Y_n}$, and optimizes the parameters by minimizing a combined loss function. This function includes the old task loss $L_{\text{old}}(Y_o, \hat{Y_o})$, the new task loss $L_{\text{new}}(Y_n, \hat{Y_n})$. 
    \item \textbf{Similarity-Preserving Distillation (SPD)}: The key idea is to ensure that both the teacher and student networks produce similar activations for the same sample, ensuring that the student model learns not only the output predictions but also the relational structure captured by the teacher model. During training, SPD enforces that the feature activation distribution in the student model matches that of the teacher model, thereby preserving the intrinsic data distribution.
\item \textbf{Continual Representation Learning (CRL)}: The input is processed by both old and new models to obtain feature activations, which are used to compute the distributions. Then, the KL divergence $D_{KL}(p_i||q_i)$ is calculated between old and new class distributions. Finally, consistency relaxation applies a margin $\delta_i$ to relax the KL divergence, resulting in the modified divergence $\mathcal{D}_{KL}'(\mathbf{p}_i||\mathbf{q}_i) = [\mathcal{D}_{KL}(\mathbf{p}_i||\mathbf{q}_i) - \delta_i]_+$.This yields the distillation loss $L^{old}$, helping the new model learn while retaining old knowledge.
\item \textbf{Adaptive Knowledge Accumulation (AKA)}: The principle of Adaptive Knowledge Accumulation (AKA) lies in dynamically adjusting the importance of various tasks or knowledge components throughout the learning process, ensuring that the model selectively retains and integrates critical information. AKA incorporates a memory bank that stores pivotal knowleges from prior tasks and links this stored knowledge to the current learning, enabling the model to reference these past experiences when encountering new tasks. This approach effectively mitigates catastrophic forgetting. In contrast to GPAK, as discussed in this paper, AKA is more suited for recognizing RGB images with higher global information entropy. However, its performance is constrained in gait sequence recognition tasks, which emphasize local pattern learning and lack detailed appearance information.
\end{itemize}
\subsection{Implementation Details}
For all protocols,the sample $x$ passes through the features extractor, generating 512-dimensional features. The features are divided into $16$ parts and inputted into GPAK, where $N^r$  is set to $64$. We randomly select $16$ identities in each iteration, each with eight silhouette sequence samples. The sequence resolution is $64 \times 44$, and the sequence length is $30$. We employ an Adam\cite{kingma2014adam} optimizer with a learning rate of $3.5 \times 10^{-4}$  and adjust the learning rate using MultiStepLR, decaying by a factor of $0.1$ in the first three steps. The trade-off factor for the loss function is set to $1.0$.
Furthermore, the same dataset is trained in equal iterations across various methods in comparative experiments. No additional data augmentation is applied to ensure fairness in evaluation. The reported results are the average of $5$ repeated experiments.

\subsection{Short-term Cross-domains Evaluation}
In this section, we evaluated the performance of GaitAdapter in Short-term cross-domain scenarios using subject-independent configuration. The cross-domain protocol primarily focuses on the model’s ability to prevent forgetting old knowledge, as shown in Table \ref{t2}. We evaluated pairwise on four datasets in the visible domain: CASIA-E (CE), OUMVLP (OU), CASIA-B (CB), and Gait3D (GD). We mainly recorded the recognition performance of the gait model on the old domain in the new scenario. Additionally, to prevent the model from overfitting on smaller domains, we also evaluated the global accuracy of two domains. The results show that our method has achieved superiority, even when the training environment changes from constrained to unconstrained conditions, such as CE changing to GD, the global accuracy still increased by 3.51$\%$ compared to the AKA method.

\begin{table*}[ht]

\caption{Short-term Cross-domain evaluation results. '$\to$' indicates the direction of domain change. 'source' represents the retroactive accuracy ($\%$) of the source domain after domain change, and 'target' represents the global accuracy ($\%$) of the new domain and the old domain}

\resizebox{\textwidth}{20mm}{
\begin{tabular}{c|cc|cc|cc|cc|cc|cc}
\hline
\multirow{2}{*}{Method} & \multicolumn{2}{c|}{CE $\to$ OU} & \multicolumn{2}{c|}{CE $\to$ GD} & \multicolumn{2}{c|}{CE $\to$ CB} & \multicolumn{2}{c|}{OU $\to$ CB} & \multicolumn{2}{c|}{GD $\to$ CB} & \multicolumn{2}{c}{OU $\to$ GD} \\ \cline{2-13} 
                        & source                & target               & source                & target               & source                & target               & source                & target               & source                & target                & source                & target               \\ \hline
SFT                     & 29.48                 & 48.77                & 41.93                 & 41.93                & 32.70                  & 34.37                & 22.27                 & 26.00                & 11.59                 & 44.77                 & 33.27                 & 33.60                 \\
LwF                     & 18.74                 & 41.22                & 19.00                 & 18.99                & 11.56                 & 13.48                & 12.14                 & 15.66                & 8.39                  & 38.74                 & 14.32                 & 14.50                 \\
SPD                     & 20.60                 & 46.47                & 23.67                 & 23.77                & 13.32                 & 15.32                & 13.79                 & 17.41                & 8.20                  & 42.51                 & 19.82                 & 20.11                \\
CRL                     & 18.57                 & 41.65                & 18.95                 & 18.94                & 11.51                 & 13.45                & 12.26                 & 15.77                & 7.99                  & 38.26                 & 14.52                 & 14.77                \\
AKA                     & 28.67                 & 49.72                & 44.43                 & 44.57                & 29.80                 & 31.45                & 45.20                 & 46.94                & 11.80                 & 45.10                 & 53.76                 & 52.71                \\ \hline
GaitAdapter           & \textbf{33.16}        & \textbf{52.63}       & \textbf{48.12}        & \textbf{48.08}       & \textbf{45.79}        & \textbf{47.40}       & \textbf{51.69}        & \textbf{53.88}       & \textbf{31.70}        & \textbf{58.72}        & \textbf{56.65}        & \textbf{55.77}       \\ \hline
\end{tabular}}
\label{t2}
\end{table*}

\subsection{Long-term Cross-domain Evaluation}

As shown in Table \ref{t3}, the experimental results of the long-term cross-domain evaluation demonstrate the effectiveness of GaitAdapter in both subject-independent and subject-dependent configurations. Under the subject-independent setting, where gallery data is excluded from training, achieving an average accuracy of 46.54\% across all datasets, this represents a significant 8.48\% point improvement over AKA, the strongest baseline method in our comparison. Notably, it maintains robust performance on CASIA-E (50.80\% mean accuracy), highlighting its ability to learn transferable representations without identity-specific fine-tuning. In contrast, the subject-dependent protocol, which incorporates gallery samples during training, reveals that GaitAdapter significantly outperforms traditional continual learning methods, with an average accuracy of 58.20\%. This represents a 15.08\% gain over AKA, particularly excelling in closed-set scenarios such as CASIA-E (51.80\%), where identity-specific learning proves advantageous.

However, the results also underscore the challenges of catastrophic forgetting in continual learning, particularly in dynamic multi-domain settings. While GaitAdapter demonstrates strong performance when gallery data is available, traditional methods (e.g., LwF, SPD) suffer severe performance degradation, indicating their inability to retain past knowledge effectively. These findings validate that GaitAdapter’s methods mitigates forgetting more effectively than existing approaches, while still maintaining robustness even when gallery participation is reduced.

\begin{table*}[]
\caption{The rank-1($\%$) of Long-term evaluation results. The training order is: CE $\to$ OU $\to$ GD $\to$ CB.}
\centering
\resizebox{\textwidth}{25mm}{
\begin{tabular}{c|cccccccccccc}
\hline
\multirow{2}{*}{\begin{tabular}[c]{@{}c@{}}Evaluation\\ Configuration\end{tabular}} & \multicolumn{1}{c|}{\multirow{2}{*}{Method}} & \multicolumn{4}{c|}{CASIA-E}                                                 & \multicolumn{1}{c|}{OUMVLP} & \multicolumn{1}{c|}{Gait3D} & \multicolumn{4}{c|}{CASIA-B}                                               & \multirow{2}{*}{avg} \\
                                                                                    & \multicolumn{1}{c|}{}                        & NM             & CL             & BG             & \multicolumn{1}{c|}{mean} & \multicolumn{1}{c|}{NM}     & \multicolumn{1}{c|}{-}      & NM           & CL             & BG             & \multicolumn{1}{c|}{mean} &                      \\ \hline
\multirow{3}{*}{\begin{tabular}[c]{@{}c@{}}Subject\\ Independent\end{tabular}}      & SFT                                          & 34.55          & 25.70          & 14.54          & 24.93                     & 17.22                       & 14.39                       & 92.70        & 87.55          & 80.54          & 86.93                     & 22.77                \\
                                                                                    & AKA                                          & 54.44          & 41.56          & 22.02          & 39.34                     & 34.86                       & 22.49                       & 94.55        & 92.60          & 83.93          & \textbf{90.36}                    & 38.06                \\ \cline{2-13} 
                                                                                    & GaitAdapter                                  & \textbf{63.52}         & \textbf{50.59}          & \textbf{24.67}         & \textbf{50.80}                     & \textbf{39.90}                       & \textbf{29.80}                       & \textbf{97.97}        & \textbf{94.30}          & \textbf{86.58}          & {87.73}                     & \textbf{46.54}                \\ \hline
\multirow{6}{*}{\begin{tabular}[c]{@{}c@{}}Subject\\ Dependent\end{tabular}}        & SFT                                          & 36.04          & 27.13          & 16.46          & 26.54                     & 20.33                       & 15.00                       & 99.99        & 99.53          & 89.5           & 96.34                     & 25.11                \\
                                                                                    & LwF                                          & 32.95          & 25.23          & 14.44          & 24.20                     & 21.22                       & 16.09                       & 99.95        & 99.16          & 89.59          & 96.23                     & 24.21                \\
                                                                                    & SPD                                          & 31.58          & 23.78          & 13.84          & 23.06                     & 20.06                       & 15.30                       & 100          & 99.63          & 91.61          & 97.08                     & 23.11                \\
                                                                                    & CRL                                          & 32.25          & 23.20          & 14.68          & 22.71                     & 21.02                       & 15.20                       & 99.8         & 98.85          & 88.24          & 95.63                     & 23.27                \\
                                                                                    & AKA                                          & 56.29          & 44.33          & 23.65          & 41.42                     & 43.96                       & 27.09                       & 100          & 99.46          & 91.73          & 97.06                     & 43.12                \\ \cline{2-13} 
                                                                                    & GaitAdapter                                  & \textbf{65.39} & \textbf{52.69} & \textbf{26.22} & \textbf{51.80}            & \textbf{66.30}              & \textbf{40.70}              & \textbf{100} & \textbf{99.65} & \textbf{96.02} & \textbf{98.55}            & \textbf{58.20}       \\ \hline
\end{tabular}}
\label{t3}
\end{table*}

\subsection{Inner-domain Evaluation}
\begin{figure}[h]
  \centering
   \includegraphics[width=75mm]{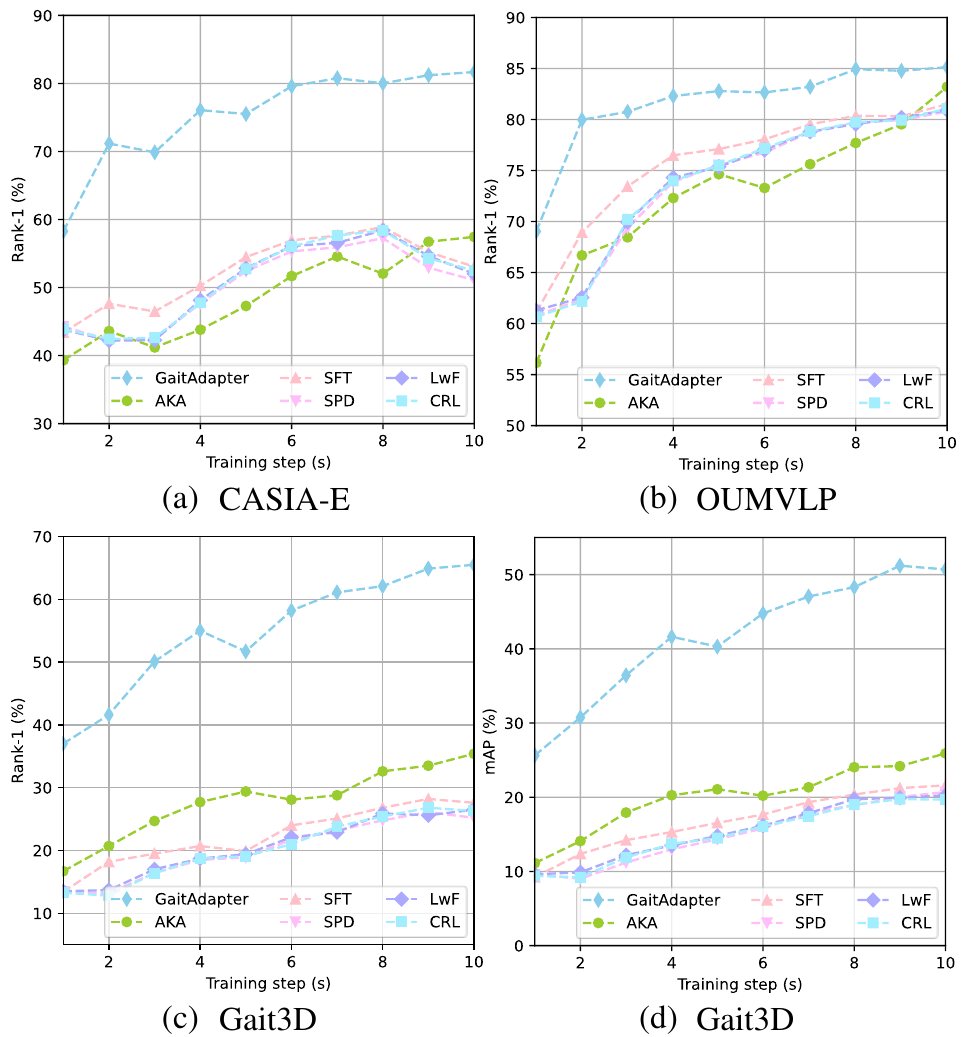}
  \caption{The Results of Inner-domain Evaluation.(a) is Rank-1(\%) on CASIA-E Datasets.(b) and (c) respectively report Rank-1 (\%) on OUMVLP and Gait3D, while (d) reports mAP (\%) on Gait3D.
}
\vspace*{-1em}
\label{f4}
\end{figure}

As shown in the figure \ref{f4}, we conducted 10-step training on the CASIA-E, OUMVLP, and Gait3D datasets, demonstrating that GaitAdapter outperforms mainstream continual learning methods, particularly on Gait3D with improvements of 29.1/37.9 (mAP/Rank-1 accuracy) over SFT and 30.09/8.89 over AKA. GaitAdapter’s adaptability to complex domains is evident, although on OUMVLP, it shows slower progress due to fewer gait patterns. Label regularization-based methods like LwF, SPD, and CRL significantly degrade gait model performance, highlighting their unsuitability for gait recognition tasks compared to GaitAdapter's approach.

\subsection{Unseen-domains Evaluation}
We used the GREW, SUStech-1K, and CCPG datasets as benchmarks for unseen-domain evaluation because, compared to visible-domain datasets, these datasets contain additional complex information that validates the model’s ability to adapt to data distribution drift. To obtain evaluable models, we sequentially trained the model on CASIA-E (CE) $\to$ OUMVLP (OU) $\to$ Gait3D (GD) $\to$ CASIA-B (CB) and only used the test set of the unseen domain for cross-domain recognition to obtain Rank-1 ($\%$) reports. The results show that our method consistently improves on all three datasets. For example, on the CCPG dataset, our method achieves an average precision improvement of 17.41$\%$ compared to SFT and 5.39$\%$ compared to AKA. The testing performance of other traditional continual methods is mostly on par with the SFT method and does not demonstrate the ability to resist changes in data distribution. The improved generalization ability of GaitAdapter in the unseen domain also demonstrates our method’s capability to capture general gait knowledge. The result is shown in Table \ref{t4}.

\begin{table}[]
\caption{The rank-1($\%$) of Unseen-domain evaluation results. BG, DN,UP,CL represent differents types of gait information.}
\centering
\resizebox{90mm}{!}{
\begin{tabular}{ccccccc|c}
\hline
\multicolumn{2}{c}{Unseen Datasets} & SFT   & LwF   & SPD   & CRL   & AKA   & GaitAdapter \\ \hline
\multicolumn{2}{c}{GREW}            & 15.76 & 15.55 & 15.38 & 15.55 & 20.95 & \textbf{21.40}   \\ \hline
\multicolumn{2}{c}{SUStech-1K}      & 11.99 & 15.48 & 12.12 & 15.22 & 16.28 & \textbf{20.24}   \\ \hline
\multirow{4}{*}{CCPG}      & BG     & 38.95 & 40.20  & 39.93 & 38.95 & 61.41 & \textbf{68.89}   \\
                           & DN     & 27.87 & 28.60  & 29.24 & 27.87 & 42.94 & \textbf{48.27}    \\
                           & UP     & 25.15 & 23.52 & 24.61 & 25.15 & 30.08 & \textbf{34.84}   \\
                           & CL     & 15.31 & 15.88 & 16.30  & 15.31 & 20.96 & \textbf{24.93}   \\ \hline
\end{tabular}
}
\label{t4}
\end{table}
\subsection{Ablation Study}
\textbf{Module Effectiveness Analysis.}The results in Table \ref{t5} highlight a consistent performance improvement across all module configurations, with the most substantial gains observed when GPAK and EDSN are combined. The inclusion of GPAK, especially with partitioning, significantly boosts accuracy, emphasizing the critical role of local pattern learning in GaitAdapt tasks. The full integration of these modules achieves the highest performance, reflecting the complementary strengths of GPAK and EDSN in enhancing generalization.

\textbf{Graph Type Analysis.}Our experimental results shown in Table \ref{t6} demonstrate that the bipartite graph structure adopted in our method effectively addresses the catastrophic forgetting problem in continual gait recognition, while maintaining competitive performance across domains. As shown in the ablation study, the bipartite graph configuration achieves superior performance on earlier domains (50.80\% on CE) compared to the fully-connected counterpart (46.26\%), indicating its stronger capability in preserving previously learned knowledge. Although the performance on later domains shows a slight decrease (87.73\% vs 92.95\% on CB), this trade-off is justified by the method's enhanced resistance to forgetting, which is crucial for long-term deployment scenarios. The bipartite design, where feature nodes only connect to meta nodes without intra-subgraph connections, creates an effective isolation mechanism that prevents interference between tasks while still allowing necessary knowledge transfer through the meta nodes. This architectural choice is particularly suitable for real-world applications where model stability is prioritized, as it significantly reduces performance degradation over time while maintaining acceptable accuracy on new tasks.

\begin{table}[]
\caption{Module Effectiveness Analysis Under Cross-Domain Evaluation Protocol. 'source' represents the retroactive accuracy ($\%$) of the source domain after domain change, and 'target' represents the global accuracy ($\%$) of the new domain and the old domain.The results represent the average performance across all cross-domain experiments.}
\centering
\resizebox{60mm}{!}{
\begin{tabular}{c|ll}
\hline
Setting                                 & \multicolumn{1}{c}{source} & \multicolumn{1}{c}{target} \\ \hline
Base                                    & 28.60                       & 38.24                      \\
Base+GPAK(w\textbackslash{}o partation) & 35.61                      & 45.08                      \\
Base+GPAK                               & 38.78                      & 48.76                      \\
Base+EDSN                               & 38.40                       & 47.79                      \\ \hline
Base+GPAK+EDSN (Full)                    & \textbf{44.51}             & \textbf{52.75}             \\ \hline
\end{tabular}}
\label{t5}
\end{table}

\begin{table}[]
\caption{The performance (Rank-1 \%) of different graph structures on long-term tasks}
\centering
\resizebox{60mm}{!}{
\begin{tabular}{ccccc}
\hline
Graph Type     & CE    & OU    & GD    & CB    \\ \hline
bipartite      & 50.80 & 39.9  & 29.8  & 87.73 \\
Full-Connected & 46.26 & 37.45 & 28.20 & 92.95 \\ \hline
\end{tabular}}
\label{t6}
\end{table}

\textbf{Impact of Partition in GPAK.}Table \ref{t7} presents the Rank-1 accuracy results under various partitioning configurations, averaged across all datasets following training. The evaluation indicates a general trend where increasing the number of partitions enhances model performance.

For instance, Rank-1 accuracy increases from 53.65\% with a single partition to 57.59\% with eight partitions. The highest performance is observed with the full partition setting (16 partitions), achieving 58.20\%. This suggests that finer partitioning enables the model to more effectively capture local patterns, which is particularly important on the GaitAdapt.

\begin{table}[]
\caption{Long-term Evaluation Results for Different Partition. The training order is: CE $\to$ OU $\to$ GD $\to$ CB. Rank-1(\%) is calculated as the average of the post-training backtesting results across all datasets.}
\centering
\resizebox{80mm}{!}{
\begin{tabular}{cccccc}
\hline
Partation Setting & CE             & OU             & GD             & CB             & Rank-1(\%)     \\ \hline
1                 & 45.48          & 63.94          & 36.70          & 98.09          & 53.65          \\
2                 & 47.55          & 61.41          & 37.31          & 98.27          & 53.80          \\
4                 & 47.82          & 62.46          & 38.60          & 98.18          & 54.41          \\
8                 & 51.25          & 65.48          & \textbf{41.30} & 98.65          & 57.59          \\ \hline
16 (Full)         & \textbf{51.80} & \textbf{66.30} & 40.70          & \textbf{98.55} & \textbf{58.20} \\ \hline
\end{tabular}}
\label{t7}
\end{table}

\section{Conclusion}

The GaitAdapt task proposed in this study opens up a new perspective in gait recognition technology. The introduction of GaitAdapter effectively addresses the critical challenges of gait recognition in continual learning scenarios. This research aims to enhance the adaptability of gait models to new identities and environments, reduce catastrophic forgetting of gait knowledge, and improve the generalization of models. Additionally, this study proposes an evaluation protocol for continual gait recognition tasks to evaluate the performance of continual gait methods. Extensive evaluations confirm that GaitAdapter can effectively manage cross-domain and cross-task gait data. It significantly outperforms other continual learning methods in various continual learning scenarios, injecting new vitality into the research of gait recognition.







\end{document}